\documentclass[lettersize,journal]{IEEEtran}
\usepackage{amsmath,amsfonts}
\usepackage[ruled,vlined,linesnumbered]{algorithm2e}
\usepackage{array}
\usepackage[caption=false,font=normalsize,labelfont=sf,textfont=sf]{subfig}
\usepackage{adjustbox}
\usepackage{multirow}
\usepackage{graphicx}
\usepackage{amssymb}

\usepackage{bigstrut}
\usepackage{textcomp}
\usepackage{stfloats}
\usepackage{url}
\usepackage{verbatim}
\usepackage{graphicx}
\usepackage{cite}
\usepackage[colorlinks,
            urlcolor=blue,
            linkcolor=blue,
            anchorcolor=blue,
            ]{hyperref}
\usepackage{booktabs}
\newcommand{\eg}{e.g.}
\newcommand{\ie}{i.e.}

\begin{document}

\title{
SAM-aware Test-time Adaptation for Universal Medical Image Segmentation}
\author{Jianghao Wu, Yicheng Wu, Yutong Xie, Wenjia Bai, You Zhang, Feilong Tang, Yulong Li, \\Imran Razzak, Daniel F Schmidt, Yasmeen George
\thanks{This research was supported by The Commonwealth of Australia under the Medical Research Future Fund No. NCRI000074. (Corresponding to Yicheng Wu. e-mail: yicheng.wu@monash.edu) \\
Jianghao Wu, Yicheng Wu, Feilong Tang, Daniel F Schmidt, and Yasmeen George are with the Department of Data Science \& AI, Faculty of Information Technology, Monash University, Melbourne, VIC 3800, Australia. \\
Wenjia Bai is with the Department of Computing and Department of Brain Sciences, Imperial College London, London, SW7 2AZ, United Kingdom. \\
Yutong Xie, Yulong Li, and Imran Razzak are with Mohamed bin Zayed University of Artificial Intelligence, Abu Dhabi, United Arab Emirates.\\
You Zhang is with the Department of Radiation Oncology, UT Southwestern Medical Center, Dallas, TX 75390-9303, United States.
}}

\maketitle
\markboth{IEEE TRANSACTIONS ON IMAGE PROCESSING, VOL. XX, NO. XX, XXXX 2025}{Jianghao Wu \MakeLowercase{\textit{et al.}}: SAM-aware Test-time Adaptation for Universal Medical Image Segmentation}

\begin{abstract}
Leveraging the Segment Anything Model (SAM) for medical image segmentation remains challenging due to its limited adaptability across diverse medical domains. Although fine-tuned variants, such as MedSAM, improve performance in scenarios similar to the training modalities or organs, they may lack generalizability to unseen data. To overcome this limitation, we propose SAM-aware Test-time Adaptation (SAM-TTA), a lightweight and flexible framework that preserves SAM’s inherent generalization ability while enhancing segmentation accuracy for medical images.
SAM-TTA tackles two major challenges: (1) input-level discrepancy caused by channel mismatches between natural and medical images, and (2) semantic-level discrepancy due to different object characteristics in natural versus medical images (e.g., with clear boundaries vs. ambiguous structures). To this end, we introduce two complementary components: a self-adaptive Bézier Curve-based Transformation (SBCT), which maps single-channel medical images into SAM-compatible three-channel images via a few learnable parameters to be optimized at test time; and IoU-guided Multi-scale Adaptation (IMA), which leverages SAM’s intrinsic IoU scores to enforce high output confidence, dual-scale prediction consistency, and intermediate feature consistency, to improve semantic-level alignments.
Extensive experiments on eight public medical image segmentation tasks, covering six grayscale and two color (endoscopic) tasks, demonstrate that SAM-TTA consistently outperforms state-of-the-art test-time adaptation methods. Notably, on six grayscale datasets, SAM-TTA even surpasses fully fine-tuned models, achieving significant Dice improvements (i.e., average 4.8\% and 7.4\% gains over MedSAM and SAM-Med2D) and establishing a new paradigm for universal medical image segmentation. Code is available at \url{https://github.com/JianghaoWu/SAM-TTA}.
\end{abstract}

\begin{IEEEkeywords}
Medical image segmentation, Test-time adaptation, Segment anything model
\end{IEEEkeywords}

\section{Introduction}
\label{sec:intro}
\begin{figure*}
    \centering
    \includegraphics[width=1\linewidth]{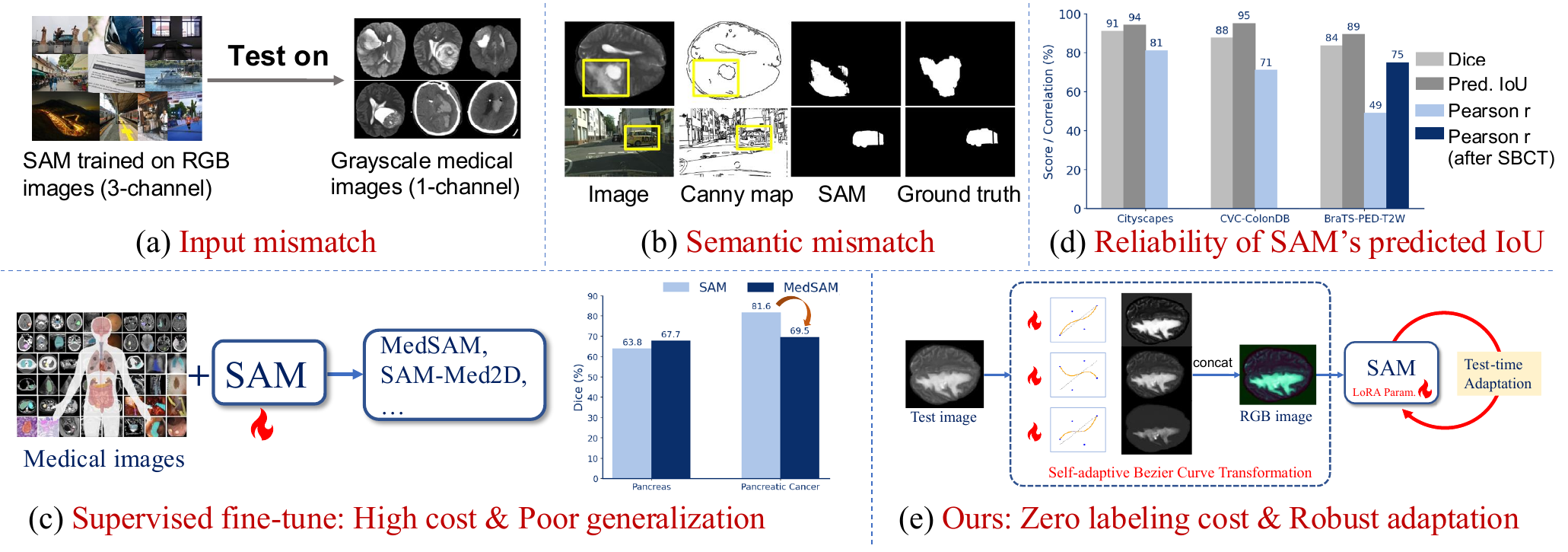}
    \caption{\textbf{Motivation of our SAM-TTA.} (a) SAM is pretrained on RGB (3-channel), while most medical images are grayscale (1-channel). (b) Anatomical targets exhibit ambiguous, low-contrast boundaries and different semantics from natural images, leading to over-/under-segmentation. (c) Supervised Fine-tuning-based approaches (e.g., MedSAM) require costly retraining and show poor generalization. (d) Dice score, predicted IoU by SAM (Pred. IoU), and their Pearson correlation $r$ on the Cityscapes, Polyp-CVC-ColonDB, and BraTS-PED-T2W datasets. (e) SAM-TTA introduces self-adaptive Bézier curve-based transformation and IoU-guided multi-scale adaptation, enabling label-free and robust test-time adaptation for universal medical image segmentation.
    }
    \label{fig:concept}
    \vskip -0.4cm
\end{figure*} 

The Segment Anything Model (SAM)~\cite{kirillov2023segment} has achieved remarkable progress across a wide range of image segmentation tasks~\cite{chen2023sam,huang2024segment}. Trained on the large-scale and diverse SA-1B dataset, SAM exhibits strong zero-shot generalization, enabling effective segmentation of previously unseen objects and domains~\cite{zhang2023segment}. This ability positions SAM as a powerful and versatile foundation model with broad downstream applications~\cite{zhang2023comprehensive}, including medical image segmentation~\cite{zhang2024segment}, where accurate delineation is critical for clinical decision-making. However, current SAM-based approaches still fall short of the accuracy and robustness required for reliable clinical deployment~\cite{wu2023medical}.

A key obstacle lies in discrepancies at both the \textit{input} and \textit{semantic} levels~\cite{zhang2024segment}, as illustrated in Fig.~\ref{fig:concept}(a)–(b). 
At the input level (Fig.~\ref{fig:concept}a), differences in acquisition protocols, imaging devices, and patient populations introduce substantial distribution shifts between natural and clinical images. 
A representative case is the channel mismatch: SAM expects three-channel RGB inputs, whereas most medical scans (e.g., CT, MRI) are single-channel grayscale images, causing direct inference failures. 
At the semantic level (Fig.~\ref{fig:concept}b), medical targets exhibit ambiguous, low-contrast boundaries and domain-specific contextual structures (e.g., organs, tumors)~\cite{zhu2023brain,lei2024condseg,Wu_2024_CVPR}, unlike natural images where object edges are typically sharp and well-defined. 
Without effective adaptation, SAM struggles to handle these input and semantic discrepancies, leading to unsatisfactory segmentation and limited reliability in downstream medical applications.

As shown in Fig.~\ref{fig:concept}(c), fine-tuning SAM appears to be a straightforward solution. By adapting SAM to large-scale medical datasets (\eg, MedSAM~\cite{ma2024segment}), it can better recognize medical structures and achieve strong results in specific domains. However, these approaches encounter two major challenges:
\textbf{High computational cost}: fine-tuning foundation models demands substantial resources and large quantities of high-quality annotations.
\textbf{Reduced generalization}: specializing SAM for a particular dataset often leads to overfitting, degrading performance on unseen domains, especially under heterogeneous clinical conditions~\cite{li2024empirical,cheng2024interactive}.
Moreover, with the rapid emergence of new imaging techniques, it is impractical to fine-tune models for all possible data distributions, further constraining SAM-based medical segmentation.

As a result, \textit{test-time adaptation} (TTA) offers a practical and efficient alternative that enables SAM to adjust to unseen data without additional labels or costly retraining. Unlike fine-tuning approaches, TTA updates only a few parameters and leverages self-supervised signals (\eg, entropy, uncertainty) to capture the unique characteristics of each test sample.
Motivated by this, we propose the \textit{SAM-aware Test-time Adaptation (SAM-TTA)} framework (see Fig.~\ref{fig:concept}(e)), which enables lightweight, label-free adaptation of SAM to medical images through efficient test-time parameter updates, providing target-domain adjustment without retraining or access to source data.
Specifically:
1) To mitigate input-level discrepancies, we introduce a \textit{Self-adaptive Bézier Curve-based Transformation (SBCT)}, which converts single-channel medical scans into SAM-compatible three-channel inputs. Unlike generative approaches (\eg, GANs~\cite{chen2022generative}, diffusion models~\cite{kazerouni2023diffusion}), SBCT requires no offline training and optimizes only a few parameters (12 variables in our implementation) at test time.
2) To address semantic-level mismatches, we design an \textit{IoU-guided Multi-scale Adaptation (IMA)} strategy that exploits SAM’s intrinsic IoU predictions as confidence signals, enforcing consistency across low- and high-resolution outputs as well as intermediate embeddings. Within a teacher–student framework, IMA reduces uncertainty and stabilizes adaptation across confidence, output, and feature spaces during inference.

Our major contributions can be summarized as follows.
\begin{itemize}
\item We propose \textbf{SAM-TTA}, a label-free, efficient test-time adaptation framework that adjusts SAM to target medical domains via lightweight parameter updates, without retraining or access to source data.
\item We design \textbf{SBCT}, a lightweight transformation that optimizes only 12 variables at test time to convert single-channel scans into structurally faithful, three-channel SAM-compatible inputs.  
\item We introduce \textbf{IMA}, which exploits SAM’s intrinsic IoU prediction to enforce feature-, output-, and confidence-level adaptation, effectively addressing semantic-level discrepancies.
\item We conduct extensive evaluations on eight public medical image segmentation tasks, covering six grayscale and two color (endoscopic) datasets, and demonstrate that SAM-TTA consistently outperforms state-of-the-art test-time adaptation approaches and even surpasses fully fine-tuned models, establishing a new paradigm for universal medical image segmentation.
\end{itemize}

\section{Related Work}

\subsection{Foundation Models for Image Segmentation}
Recent advances in vision foundation models, such as CLIP~\cite{radford2021learning}, DINO~\cite{oquab2023dinov2}, and particularly the Segment Anything Model (SAM)~\cite{kirillov2023segment}, have unlocked new possibilities for generalizable image segmentation~\cite{awais2025foundation,zhang2024challenges}. Trained on large-scale data, SAM provides powerful zero-shot segmentation via flexible prompts (points, boxes, masks, and text). Its success has stimulated growing interest in extending SAM to medical image segmentation~\cite{cheng2023sam_li}.
Several adaptations have been proposed. MedSAM~\cite{ma2024segment} fine-tunes SAM for medical images while retaining its box-prompt functionality. SAM-Med2D~\cite{cheng2023sam} adopts adapter-based fine-tuning on 4.6 million medical images, freezing the image encoder, inserting lightweight adapters into each Transformer block, and updating the prompt encoder and mask decoder. For volumetric data, MA-SAM~\cite{chen2024ma} introduces modality-agnostic 3D adapters into the image encoder, enabling segmentation across CT, MRI, and surgical videos by incorporating volumetric and temporal cues. MedCLIP-SAM~\cite{koleilat2024medclip} further leverages CLIP’s text encoder for text-driven segmentation tasks such as delineating the left ventricle.
Despite their success, most SAM-based adaptations depend heavily on large-scale annotations or extensive fine-tuning, which is impractical under significant domain shifts. Meanwhile, considering medical imaging is rapidly evolving, our work explores unsupervised test-time adaptation, aiming to reduce reliance on annotated data while fully exploiting the pretraining generalization capacity of SAM.

\subsection{Test-Time Adaptation}  
TTA adapts pretrained models using only unlabeled target data at inference, without requiring access to the source training set or performing large-scale retraining~\cite{Wen_2024_WACV,pmlr-v227-valanarasu24a,yuan2023robust,yang2024test}. Unlike conventional fine-tuning, TTA must update models on-the-fly, often within a single pass~\cite{chen2024each,liang2025comprehensive}.
Early work explored auxiliary self-supervised tasks. Sun et al.~\cite{sun2020test} introduced a rotation-prediction branch, while Karani et al.~\cite{karani2021test} proposed shallow normalization networks optimized with a denoising auto-encoder. However, such modifications to the segmentation model are less suitable for large pretrained FMs such as SAM~\cite{kirillov2023segment}.
More recent methods operate in the fully test-time setting, without auxiliary supervision or source data. Batch-normalization-based methods such as PTBN~\cite{nado2020evaluating} and TENT~\cite{wang2021tent} adapt feature statistics, while WCEM~\cite{lee2023towards} and SAR~\cite{wang2023dynamically} improve stability by filtering unreliable predictions. Self-training approaches, such as UPL-TTA~\cite{wu2023upltta}, generate pseudo-labels using weak augmentations, dropout, and multi-decoder ensembles. CoTTA~\cite{wang2022continual} stabilizes continual adaptation via teacher updates with stochastic weight averaging and pseudo-label consistency. GraTa~\cite{chen2025gradient} aligns pseudo gradients with entropy-derived gradients and employs cosine-similarity learning rates for more stable optimization. Recently, Chen et al.~\cite{kecheng2025ICCV} refined MedSAM at test time by updating only the image embedding with a latent CRF loss plus entropy minimization, avoiding parametric updates.
Most existing approaches focus primarily on prediction-level refinement, keeping the input representation fixed. However, when the target domain shift arises mainly from intensity and contrast differences, such prediction-only strategies provide limited alignment with the pretrained distribution, leading to limited performance. In contrast, our approach explicitly introduces input-level adaptation via an adaptive input transformation, which better aligns medical images with SAM’s learned features. Combined with IoU-guided semantic alignment, this enables more effective test-time adaptation across diverse medical domains, without altering SAM’s architecture.

\section{Method}\label{sec:method}
\begin{figure*}
    \centering
    \includegraphics[width=0.98\linewidth]{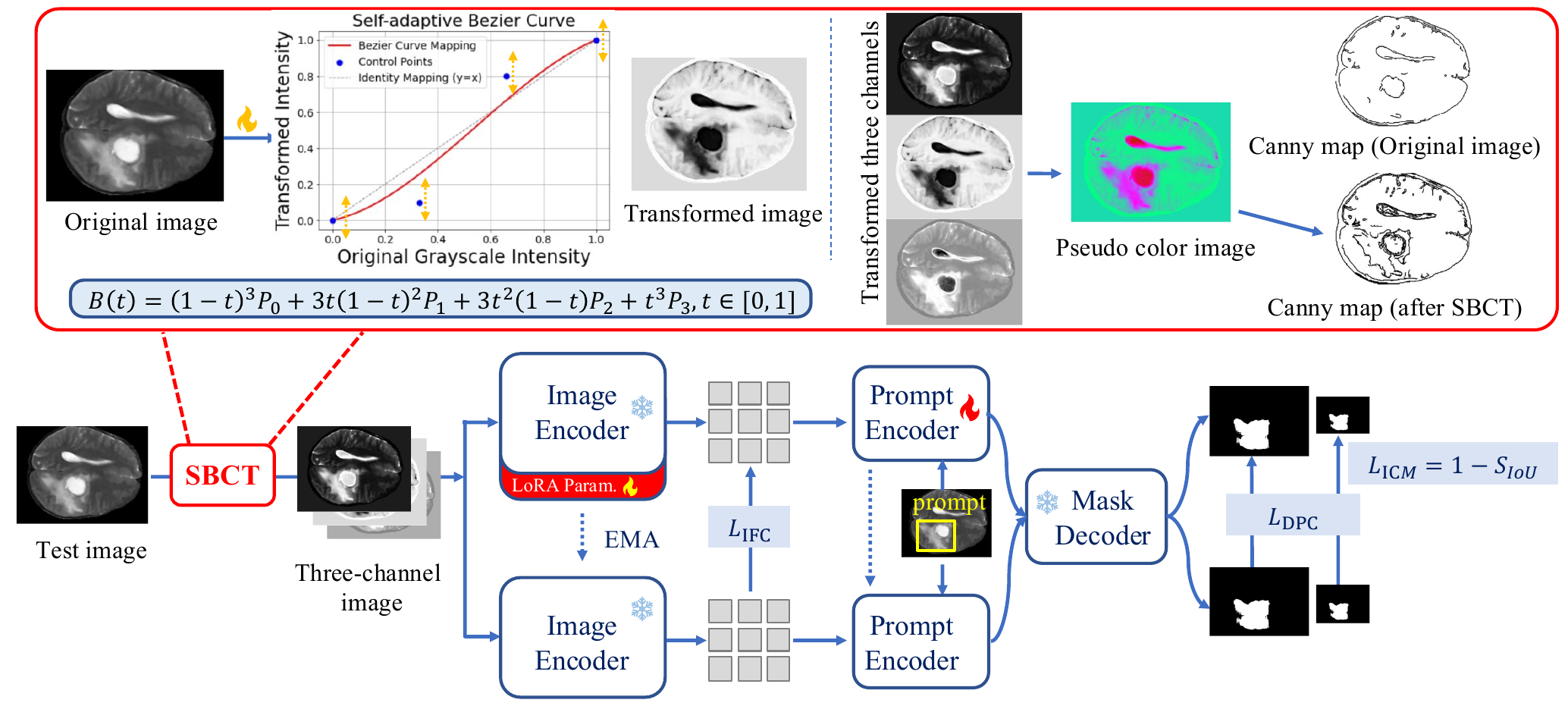}
\caption{\textbf{Overview of the proposed SAM-TTA framework.} 
At test time, a grayscale medical image is transformed by \emph{Self-adaptive Bézier Curve-based Transformation} (SBCT) into an image with three channels to be compatible with SAM input. The transformed image is then segmented by SAM to generate low- and high-resolution masks with an intrinsic IoU score.
Model optimization is guided by the proposed \emph{IoU-guided Multi-scale Adaptation} (IMA) strategy, which consists of:
(i) \emph{IoU Confidence Maximization} ($\mathcal{L}_{\text{ICM}}$), reducing prediction uncertainty via SAM’s IoU prediction head;
(ii) \emph{Dual-scale Prediction Consistency} ($\mathcal{L}_{\mathrm{DPC}}$), enforcing agreement between student and teacher predictions across scales; and
(iii) \emph{Intermediate Feature Consistency} ($L_{\mathrm{IFC}}$), aligning encoder embeddings between student and teacher models.}
    \label{fig:pipeline}
    \vskip -0.4cm
\end{figure*}
In this work, we aim to leverage the powerful SAM model for universal medical image segmentation in the test-time adaptation scenario.  As illustrated in Fig.~\ref{fig:pipeline}, our proposed SAM-TTA has two key components:
(1) a \emph{Self-adaptive Bézier Curve-based Transformation} (SBCT), which converts single-channel medical scans into three-channel SAM-compatible inputs, aiming at alleviating input-level discrepancies; and
(2) an \emph{IoU-guided Multi-scale Adaptation} (IMA), which exploits SAM’s intrinsic IoU scores to enhance segmentation reliability by maximizing IoU predictions, enforcing dual-scale segmentation consistency, and aligning intermediate features to mitigate semantic-level discrepancies during inference.

\subsection{Overview of Segment Anything Model}
SAM, denoted by $F$, consists of three primary modules. First, the image encoder, $z = f(X)$, extracts visual representations from the input image $X$ and is pretrained using a masked autoencoder (MAE)~\cite{he2022masked}. Second, the prompt encoder, $e = g(p)$, encodes user-provided prompts $p$ to guide segmentation. Third, the mask decoder, $h(z, e)$, fuses the image and prompt features to generate segmentation masks. After pretraining the image encoder, SAM undergoes fine-tuning on the large-scale SA-1B dataset~\cite{kirillov2023segment}. During inference, the image encoder processes a test image $X$ to produce feature representations $z = f(X)$. Then, given encoded prompts $e$, the mask decoder produces three outputs: a coarse low-resolution mask $M_l$, a refined high-resolution mask $M_h$. Another lightweight MLP head is designed to produce an IoU prediction $S_{\text{IoU}}$ that estimates segmentation quality.

Here, $S_{\text{IoU}}$ is trained on SA-1B by using MSE loss:
\begin{equation}
\mathcal{L}_{\text{IoU}}=\big(S_{\text{IoU}}-\text{IOU}(M_h,Y)\big)^2,
\end{equation}
where $Y$ is the ground truth and $IOU(M_h, Y) = \frac{|M_h \cap Y|}{|M_h \cup Y|}$
denotes the real IoU score. Based on the large-scale training, 
$S_{\text{IoU}}$ serves as an intrinsic metric to measure the quality of generated masks. 
To validate the reliability of SAM’s intrinsic IoU prediction, we conducted a preliminary experiment summarized in Fig.~\ref{fig:concept}(d). The figure reports the Dice, predicted IoU (Pred. IoU), and their Pearson correlation ($r$) on one natural image dataset (Cityscapes) and two medical image datasets (Polyp-CVC-ColonDB and BraTS-PED-T2W). While Pred. IoU generally overestimates segmentation quality, it maintains a strong correlation with Dice on RGB datasets, with $r$ consistently exceeding 0.7 on Cityscapes and Polyp-CVC-ColonDB. This indicates that $S_{\text{IoU}}$ serves as a \textbf{\textit{generalized guidance signal}} for confidence estimation. In contrast, on the grayscale BraTS-PED-T2W dataset, the correlation is weaker ($r$~=~0.49) due to modality mismatch; however, after applying our proposed SBCT (detailed in the next subsection), the $r$ rises markedly to~0.75, indicating that SBCT alleviates input-level discrepancies and restores the reliability of SAM’s confidence estimation in grayscale domains.

Building on these observations, our SAM-TTA  leverages SAM’s strong representational capacity and intrinsically generalizable outputs to enable efficient adaptation to diverse medical domains under a test-time setting. We now introduce its core components, beginning with the SBCT module.

\subsection{Self-adaptive Bézier Curve-based Transformation}
Since SAM is pretrained on natural RGB images, its encoder leverages complementary luminance and color-opponent cues from three channels. In contrast, most medical modalities (e.g., CT, MRI) are single-channel grayscale. A common practice is to duplicate the grayscale image across three channels~\cite{ma2024segment}. While straightforward, this produces identical channels and fails to capture inter-channel variation, thus under-utilizing SAM’s reliance on color diversity and weakening adaptability under domain shifts. A more effective strategy is to generate diverse channel mappings $x_i = f_i(x)$ that enrich representations. Therefore, we propose a lightweight Self-adaptive Bézier Curve-based Transformation (SBCT) to achieve this. Instead of relying on auxiliary generators, SBCT reparameterizes intensities using three per-image cubic Bézier curves, one for each output channel, optimized directly at test time. This design introduces only 12 learnable scalar parameters, providing flexible channel diversification without external supervision.

Specifically, let $t\in[0,1]$ denote the normalized intensity of a pixel, and $P_c=\{P_{c,0},P_{c,1},P_{c,2},P_{c,3}\}$ are four distinct control points of the  Bézier curve for the $c_{th}$ channel ($c\in{1,2,3}$). We fix the $x$-coordinates to ${0,1/3,2/3,1}$ and learn the $y$-coordinates via sigmoid normalization:
\begin{equation}
P_{c,j}=\sigma(u_{c,j}) \in [0,1], \quad j\in{0,1,2,3},
\end{equation}
where ${u_{c,j}}$ are trainable scalars. Hence, the cubic Bézier curve becomes:
\begin{equation}
B(t;P_c)=(1-t)^3 P_{c,0}+3t(1-t)^2 P_{c,1}+3t^2(1-t)P_{c,2}+t^3 P_{c,3},
\end{equation}
For single-channel data $X\in\mathbb{R}^{H\times W}$, three desired channels can be generated by such point-wise mapping:
\begin{equation}
\hat{X}^{(c)}(h,w)=B\big(X(h,w);P_c\big), \quad c\in{1,2,3},
\end{equation}
yielding the SAM-compatible three-channel input $\hat{X}=\operatorname{concat}\big(\hat{X}^{(1)},\hat{X}^{(2)},\hat{X}^{(3)}\big)\in\mathbb{R}^{3\times H\times W}
$. For color medical data, such as endoscopic images, SBCT can also perform this transformation as $\hat{X}^{(c)}(h,w)=B\big(X^{(c)}(h,w); P_c\big), \quad c\in{1,2,3}$.

The advantages of our SBCT design can be summarized as follows:
1) Since grayscale medical scans lack real RGB counterparts, SBCT does not rely on ground-truth supervision and can be optimized end-to-end by using segmentation signals.
2) SBCT contains only 12 variables, enabling efficient optimization at test time.
3) The Bézier curve formulation enforces a global intensity transformation, effectively preventing local distortions or anatomical hallucinations~\cite{wu2024fpl-plus}.

\subsection{IoU-guided Multi-scale Adaptation}
The SBCT-transformed image $\hat{X}$ is fed into the SAM model to produce segmentation outputs.
Here, we propose the IoU-guided Multi-scale Adaptation (IMA) strategy to optimize the adaptation parameters through three complementary stages:

\paragraph{IoU Confidence Maximization}
Since $S_{\text{IoU}}$ serves as a measure of segmentation reliability, the first objective is to directly maximize this confidence score for each adapted embedding $\hat{z}$:
\begin{equation}
\mathcal{L}_{\text{ICM}}=1-S_{\text{IoU}},
\label{eq:ICM}
\end{equation}
which promotes highly confident segmentation even without ground-truth annotations.

To address the semantic discrepancies between medical and natural domains, we further introduce Low-Rank Adaptation (LoRA) modules within the image encoder (while keeping the mask decoder frozen to preserve the calibration of the intrinsic IoU predictor) to perform domain alignment through consistency training.
In addition, a teacher–student framework is employed, where the teacher’s parameters are updated as the Exponential Moving Average (EMA) of the student’s, to ensure stable and smooth adaptation. Following the online test-time adaptation setting, the model is adapted sequentially across the test stream, where the adapted parameters from the previous sample serve as initialization for the next one. This enables the model to accumulate domain knowledge while maintaining stability through gradual EMA updates.

\paragraph{Dual-scale Prediction Consistency} SAM naturally produces predictions at two resolutions, where the low-resolution mask captures coarse semantic structure and the high-resolution mask refines fine boundary details. 
Let $\hat{M}_h^S,\hat{M}_l^S$ and $\hat{M}_h^T,\hat{M}_l^T$ denote the student and teacher predictions at high and low resolutions, respectively. 
The dual-scale prediction consistency loss is formulated as:
\begin{equation}
\mathcal{L}_{\text{DPC}}=\mathcal{S}_{\text{Dice}}(\hat{M}_h^S,\hat{M}_h^T)+\mathcal{S}_{\text{Dice}}(\hat{M}_l^S,\hat{M}_l^T),
\label{eq:DPC}
\end{equation}
with the soft Dice function defined as:
\begin{equation}
S_{\text{Dice}}(A,B)=1-\frac{2\sum_i A_iB_i+ \epsilon}{\sum_i A_i+\sum_i B_i + \epsilon},
\end{equation}
where $\epsilon$ is a small constant to avoid zero division. Then, to reduce the guidance of unreliable samples, the loss is weighted by the normalized IoU scores:
\begin{equation}\label{equation8}
\lambda_{\text{DPC}} = \frac{-\log(1 - S_{\text{IoU}} + \epsilon)}{\max_{k \leq b}[-\log(1 - S_{\text{IoU}}^{(k)} + \epsilon)]},
\end{equation}
where $\max_{k\le b}$ is the running maximum over all images processed up to (and including) the current one $b$.
This normalizes the scale across the stream and assigns larger weights to more confident samples.

\paragraph{Intermediate Feature Consistency}
Beyond the input- and output-level alignment, the feature-level consistency encourages the student to match the teacher’s intermediate representations, mitigating representation shift.
Given the student and teacher features $\hat z^S,\hat z^T\in\mathbb{R}^{D\times H\times W}$, we normalize each channel spatially using a temperature-scaled softmax:
\begin{equation}
P_d^T=\mathrm{softmax}(\tfrac{\hat z_d^T}{\tau+\epsilon}), \quad P_d^S=\mathrm{softmax}(\tfrac{\hat z_d^S}{\tau+\epsilon}),
\end{equation}
where the temperature $\tau$ is adaptively set to the predicted $S_{\text{IoU}}$, and $d \in 1,...,D$ indexes the feature channels. Here, the spatial normalization is applied independently to each channel. These features are encouraged to align with their reliable counterparts (\ie, the feature representation of the teacher model $F^T$).
The intermediate feature-level consistency is then enforced via the KL divergence:
\begin{equation}
\mathcal{L}_{\text{IFC}}=\tfrac{1}{DHW}\sum_{d=1}^D\sum_{i=1}^{HW} P_d^T(i)\log\frac{P_d^T(i)}{P_d^S(i)}
\label{eq:IFC}
\end{equation}

Together, these three stages jointly encourage confident, consistent, and semantically aligned predictions, enabling robust test-time adaptation of SAM in diverse medical imaging scenarios without requiring any ground-truth labels.

\subsection{Overall Objective}
The final test-time optimization objective combines the three complementary components described above:
\begin{equation}
\mathcal{L}_{\text{TTA}}=\mathcal{L}_{\text{ICM}}+\lambda_\text{DPC}\times\mathcal{L}_{\text{DPC}}+\lambda_\text{IFC}\times\mathcal{L}_{\text{IFC}}.
\end{equation}
Here, $\lambda_{\text{DPC}}$ is dynamically determined by $S_{IoU}$ (Eq.~\ref{equation8}), while $\lambda_{\text{IFC}}$ is a fixed constant for balance.

\begin{algorithm}[t]
\DontPrintSemicolon
\SetAlgoLined
\SetKwInput{KwIn}{Input}
\SetKwInput{KwOut}{Output}
\SetKw{StopGrad}{stop-gradient}
\caption{SAM-aware Test-time Adaptation}
\label{alg:unsup-sam-adapt}
\KwIn{Dataset $\mathcal{D}=\{X_i\}_{i=1}^{N}$, SAM $F$, prompt $p$, optimizer, EMA decay $\alpha$, loss weight $\lambda_\text{IFC}$.}
\KwOut{Adapted student $F^S$ and its high-resolution predictions $\{\hat M_h^{(i)}\}_{i=1}^{N}$}

\textbf{Init:} $F^S \gets F,\; F^T \gets F$; running max $m \gets 0$ \;

\ForEach{$X \in \mathcal{D}$}{
  $\hat X \gets \mathrm{concat}\big(B(X;P^1),\,B(X;P^2),\,B(X;P^3)\big)$ \tcp*[r]{SBCT remap}
  $(\hat M_l^S,\,\hat M_h^S,\,S_{\mathrm{IoU}},\,\hat z^S) \gets F^S(\hat X, p)$ \;
  $(\hat M_l^T,\,\hat M_h^T,\,\hat z^T) \gets F^T(\hat X, p)$   \StopGrad
\tcp*[r]{Forward}
  $m \gets \max\!\big(m,\,-\log(1 - S_{\mathrm{IoU}} + \epsilon)\big)$;\;
  $\lambda_{\text{DPC}} \gets \dfrac{-\log(1 - S_{\mathrm{IoU}} + \epsilon)}{\,m}$;\;
  compute $\mathcal{L}_{\text{ICM}}$ via Eq.~\eqref{eq:ICM};\;
  compute $\mathcal{L}_{\text{DPC}}$ via Eq.~\eqref{eq:DPC};\;
  compute $\mathcal{L}_{\mathrm{IFC}}$ via Eq.~\eqref{eq:IFC};\;
  $\mathcal{L}_{\mathrm{TTA}} \gets \mathcal{L}_{\text{ICM}} + \lambda_{\text{DPC}}\times\mathcal{L}_{\text{DPC}} + \lambda_{\text{IFC}}\times\mathcal{L}_{\mathrm{IFC}}$;\;
\texttt{optimizer.zero\_grad()};\;
  back-propagate $\nabla \mathcal{L}_{\mathrm{TTA}}$ to update student $\theta^{S}$ and the 12 SBCT variables;\;
  $\theta \gets \alpha\,\theta + (1-\alpha)\,\theta^{S}$ \quad \text{for each } $\theta \in F^T$ \;
  $\hat M_h \gets F^S(\hat X, p)$;\; save $\hat M_h$ \;
}
\Return Adapted student $F^S$ and predictions $\{\hat M_h^{(i)}\}_{i=1}^{N}$\;
\end{algorithm}

In summary, as detailed in Algorithm~\ref{alg:unsup-sam-adapt}, for each test image $X$, the designed SBCT strategy first produces a SAM-compatible input $\hat{X}$. The IMA module then minimizes $\mathcal{L}_{\text{TTA}}$ for several gradient steps, updating the student’s LoRA adapters and the 12 SBCT scalars. The student’s high-resolution mask $\hat{M}_h$ is taken as the final prediction. This procedure is repeated sequentially across all test images.

\begin{figure*}
    \centering
    \includegraphics[width=\linewidth]{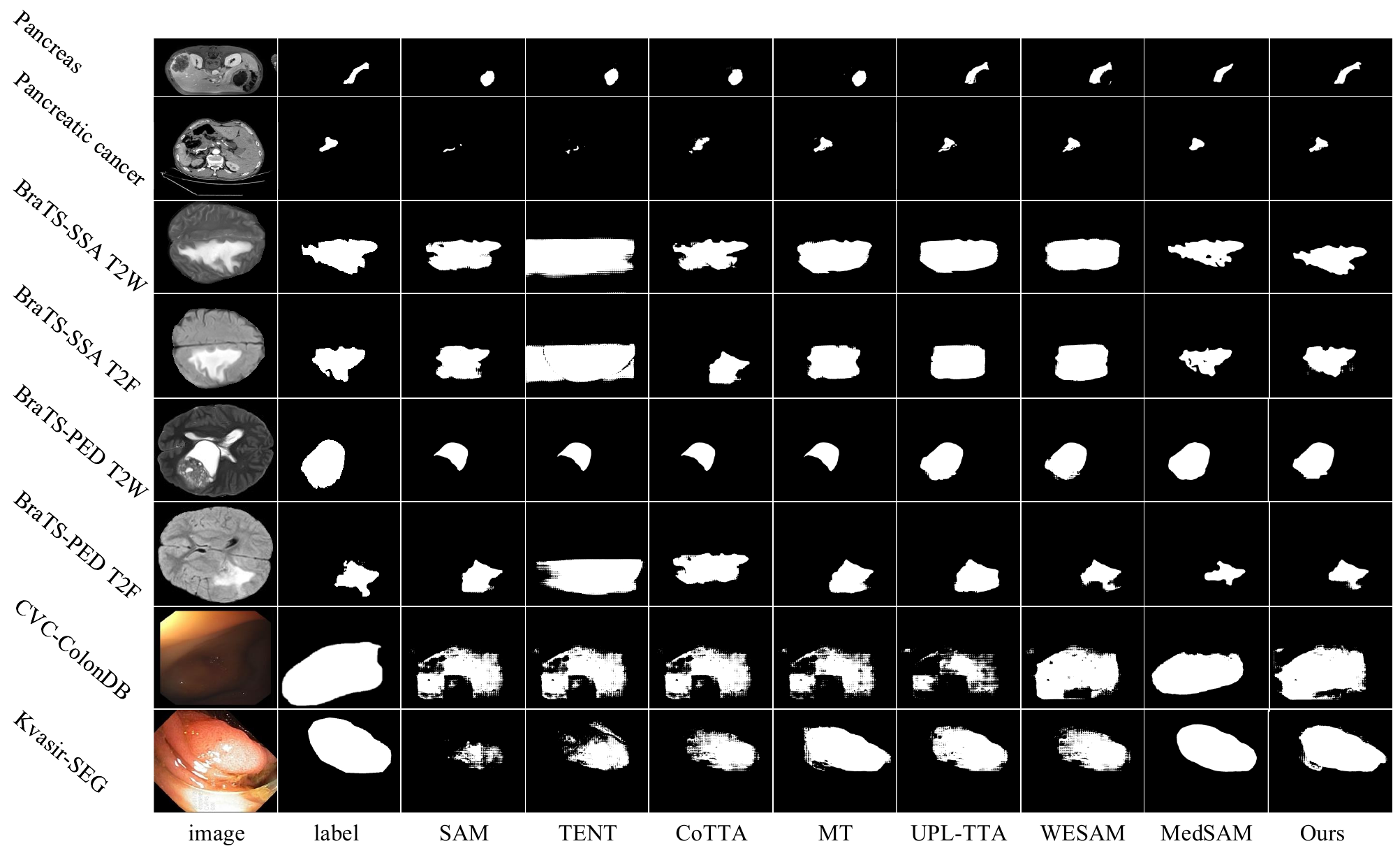}
    \caption{Visual comparison of segmentation results across multiple medical image datasets. 
    }
    \label{fig_seg_sota}
\end{figure*}

\section{Experiments and Results}

\subsection{Experimental Setup}
\subsubsection{Backbone and Prompt}
Following MedSAM~\cite{ma2024segment}, we adopt ViT-B~\cite{dosovitskiy2020image} as the image encoder and use SAM’s standard prompt and mask encoders. 
For quantitative evaluation, we follow the Bounding-box prompt protocol~\cite{ma2024segment,cheng2023sam}.
To ensure a stable and fair benchmark, the bounding boxes are oracle prompts derived from ground-truth masks and are consistently applied to all compared methods, as well as to both the student and teacher branches in our framework. 
\subsubsection{Compared Methods}
We evaluate multiple test-time adaptation approaches and one SAM-based source-free domain adaptation method. In our evaluations, direct testing of the pretrained SAM is used as the baseline (\textbf{SAM}).
\textbf{TENT}~\cite{wang2021tent} that updates the parameters by minimizing the entropy of model predictions on new test data; 
\textbf{MT}~\cite{tarvainen2017mean} that utilizes a teacher-student structure for adaptation;
\textbf{CoTTA}~\cite{wang2022continual} that employs test-time augmentation-based pseudo labels for adaptation;
\textbf{UPL-TTA}~\cite{wu2023upltta} that introduces a weak transformation and a multi-decoder to generate reliable pseudo labels and diverse predictions for adaptation;
\textbf{WESAM}~\cite{zhang2024improving} that uses a self-training architecture with anchor network regularization and contrastive loss regularization for source-free SAM adaptation.
\textbf{GraTa}~\cite{chen2025gradient} that aligns the pseudo gradient with an auxiliary entropy-derived gradient and adopts a cosine-similarity–based dynamic learning rate to improve test-time optimization for segmentation.
We also compared our method with two larger-scale medical fine-tuned SAM models: \textbf{SAM-Med2D}\cite{cheng2023sam}, which uses adapter layers to fine-tune SAM on extensive medical images, and \textbf{MedSAM}\cite{ma2024segment}, which fine-tunes SAM's image encoder and mask decoder on a large annotated dataset.

\begin{table}[h]
\centering
\caption{Datasets and the number of test images used in this work.}
\resizebox{1\linewidth}{!}{
\begin{tabular}{l|c|c}
\toprule
\textbf{Dataset} & \textbf{Modality} & \textbf{Test Samples}\\
\hline
BraTS-SSA T2W~\cite{adewole2023brain}   & T2-weighted   & 2165    \\
BraTS-SSA T2F~\cite{adewole2023brain}   & T2 FLAIR   & 2165    \\
BraTS-PED T2W~\cite{kazerooni2024brain}   & T2-weighted   & 1554    \\
BraTS-PED T2F~\cite{kazerooni2024brain}   & T2 FLAIR   & 1554    \\
Pancreas~\cite{ji2022amos}         & MRI        & 1000   \\
Pancreatic cancer~\cite{wu2025mswal}         & CT        & 1840   \\
CVC-ColonDB~\cite{bernal2015wm}        & Endoscopic  & 266   \\
Kvasir-SEG~\cite{jha2020kvasir}  & Endoscopic  & 700    \\
\bottomrule
\end{tabular}
}
\label{tab:datasets}
\end{table}

\subsubsection{Datasets} 
We evaluate the generalization capability of our method on eight diverse medical image segmentation tasks covering multiple modalities and anatomical targets, as summarized in Table~\ref{tab:datasets}: 
(1) \textbf{Brain Tumor Segmentation:} Two subsets from the BraTS2023 challenge are used for brain tumor segmentation: 
(a) \textbf{BraTS-SSA:} The Sub-Saharan Africa adult glioma dataset~\cite{adewole2023brain}, evaluated on T2-weighted (T2W) and T2-FLAIR (T2F) MRI scans; 
(b) \textbf{BraTS-PED:} The pediatric brain tumor dataset~\cite{kazerooni2024brain}, also evaluated on T2W and T2F MRI scans. 
(2) \textbf{Pancreas Segmentation:} The AMOS 2022 challenge dataset~\cite{ji2022amos} offers abdominal MRI scans with multi-organ annotations, from which we extract the pancreas labels for organ-level segmentation.  
(3) \textbf{Pancreatic Cancer Segmentation:} The MSWAL dataset~\cite{wu2025mswal} offers CT scans with voxel-level annotations of pancreatic tumors, enabling evaluation on lesion-focused segmentation. (4) \textbf{Polyp Segmentation:} 
Two endoscopic RGB benchmarks are adopted: 
(a) \textbf{CVC-ColonDB:} A colonoscopy dataset with pixel-level polyp annotations~\cite{bernal2015wm}; 
(b) \textbf{Kvasir-SEG:} Another colonoscopy dataset with diverse polyp appearances and sizes~\cite{jha2020kvasir}.

\subsubsection{Evaluation Metrics}
We report Dice Score (DSC) and 95th percentile Hausdorff Distance (HD95) as evaluation metrics. DSC measures the overlap between the predicted and ground-truth segmentation masks, while HD95 evaluates the boundary agreement by computing the 95th percentile of the Hausdorff Distance between them.

\begin{table*}[htbp]
  \centering
\caption{Quantitative comparison with other methods on grayscale medical image datasets. \emph{Note:} Pancreas is in-distribution for the MedSAM and SAM-Med2D, whereas Pancreatic cancer, BraTS-SSA, and BraTS-PED were not used to train these models.}
  \resizebox{1.0\linewidth}{!}{
    \begin{tabular}{l|cc|cc|cc|cc|cc|cc|cc}
    \toprule
    \multirow{2}[4]{*}{Method}  & \multicolumn{2}{c|}{Pancreas} & \multicolumn{2}{c|}{Pancreatic cancer}& \multicolumn{2}{c|}{BraTS-SSA T2W} & \multicolumn{2}{c|}{BraTS-SSA T2F} & \multicolumn{2}{c|}{BraTS-PED T2W} & \multicolumn{2}{c|}{BraTS-PED T2F} & \multicolumn{2}{c}{Average} \bigstrut\\
\cline{2-15}          & DSC & HD95  & DSC & HD95  & DSC & HD95  & DSC & HD95  & DSC & HD95  & DSC & HD95  & DSC & HD95 \bigstrut\\
    \hline
    SAM~\cite{kirillov2023segment}  & 63.79  & 19.60 & 81.57  & 6.12   & 78.77  & 25.06  & 79.77  & 23.49  & 83.62 & 15.71 & 85.67 & 13.38 & 78.87  & 17.23  \bigstrut[t]\\
    TENT~\cite{wang2021tent} & 63.81 &19.60  & 79.57 & 7.78   & 49.76  & 45.00 & 62.70 & 30.83  & 68.50 & 27.47 & 65.31 & 29.13 & 64.96  & 26.64  \\
    CoTTA~\cite{wang2022continual}  &64.26  &19.44  & 81.60  & 6.10 & 79.91  & 24.83  & 80.80  & 23.45  & 84.08  & 15.49  & 86.16  & 13.01  & 79.47 & 17.05   \\
    MT~\cite{tarvainen2017mean}    &65.37  &19.12  & 83.71  & 5.22  & 81.41  & 23.87  & 81.72  & 23.73  & 87.97 & 11.87 & 88.63 & 11.04  & 81.47 & 15.81  \\
    UPL-TTA~\cite{wu2023upltta}  &70.64  &15.38  & 82.76  & 7.31 & 80.10 & 26.78  & 80.28  & 26.41  & 87.74 & 12.36 & 88.78 & 10.93 & 81.72  & 16.53  \\
    WESAM~\cite{zhang2024improving}  &70.44  &16.31  & 84.25  & 5.06 & 81.75  & 22.36  & 80.63  & 25.33  & 87.63 & 12.42 & 88.35 & 11.24 & 82.18  & 15.45  \\
    GraTa~\cite{chen2025gradient} &64.45  &19.44  & 82.69  & 5.74 & 77.91  & 25.46  & 79.21  & 24.57  & 84.69 & 14.85 & 86.32 & 13.16 & 79.21  & 17.20  \\
    SAM-TTA (Ours)  & \textbf{74.41} &\textbf{14.26}   & \textbf{84.68}  & \textbf{4.92} & \textbf{83.52}  & \textbf{20.15}  & \textbf{88.66}  & \textbf{11.45}  & \textbf{89.28}  & \textbf{9.58}  & \textbf{90.98}  & \textbf{7.26}  & \textbf{85.26}  & \textbf{11.27}   \\
    \hline
    SAM-Med2D~\cite{cheng2023sam} &\underline{79.36}  &20.74   & \underline{73.68}  & 16.28& 72.85 & 32.81 & 79.32  & 30.48  & 77.09  & 27.70 & 84.87  & 23.76   & 77.86 & 25.30 \\
    MedSAM~\cite{ma2024segment} & 67.71 &\underline{15.36}   & 69.50  & \underline{11.34}  & \underline{82.70} & \underline{19.50} & \underline{85.33}  & \underline{15.86}  & \underline{87.38}  & \underline{11.70} & \underline{90.05}  & \underline{8.08}  & \underline{80.45} & \underline{13.64}  \\
    \bottomrule
    \end{tabular}%
    }
  \label{tab_sota_gray}
\end{table*}%

\begin{table}[htbp]
  \centering
  \caption{Quantitative comparison on RGB medical image datasets (CVC-ColonDB and Kvasir-SEG). Evaluation metrics include DSC (\%) and HD95 (pixels).}
  \resizebox{1\linewidth}{!}{
    \begin{tabular}{l|cc|cc}
    \toprule
    \multirow{2}[4]{*}{Method} & \multicolumn{2}{c|}{CVC-ColonDB} & \multicolumn{2}{c}{Kvasir-SEG} \bigstrut\\
\cline{2-5}          & DSC & HD95  & DSC  & HD95 \bigstrut\\
    \hline
    SAM~\cite{kirillov2023segment}   & 87.58   & 8.20 & 80.95   & 14.93 \bigstrut[t]\\
    TENT~\cite{wang2021tent}  & 87.68 & 8.10 & 82.62   & 15.58 \\
    CoTTA~\cite{wang2022continual} & 87.60  & 8.17 & 81.01  & 14.90 \\
    MT~\cite{tarvainen2017mean}    & 87.74  & 7.96  & 83.58  & 13.92 \\
    UPL-TTA~\cite{wu2023upltta} & 87.83   & 7.98 & 82.70  & 13.95 \\
    WESAM~\cite{zhang2024improving} & 87.97 & 7.95  & 82.70  & 13.95 \\
    GraTa~\cite{chen2025gradient} &87.61 &8.18 &81.11 &14.59 \\
    SAM-TTA (Ours) & \textbf{89.08}  & \textbf{7.28}  & \textbf{85.55}   & \textbf{13.15} \\
    \hline
    SAM-Med2D~\cite{cheng2023sam} & \underline{90.34} & 6.57 & \underline{92.49} & \underline{8.38} \\
    MedSAM~\cite{ma2024segment} & 90.27  & \underline{5.72}  & 89.43   & 8.49 \\
    \bottomrule
    \end{tabular}%
    }
  \label{tab_sota_RGB}%
\end{table}

\subsubsection{Implementation Details}
We optimize SBCT variables, the LoRA module of the ViT-B image encoder, and the entire prompt encoder using the Adam optimizer across all experiments. The LoRA module for the image encoder is configured with a low-rank setting of 4. The batch size is set to 1. The four control points per channel (totaling 12 scalars for three channels) use lr=0.01, enabling rapid adaptation, while the learning rate for the LoRA and prompt encoder is 0.001, with a weight decay of 0.0001. $\epsilon$ is 1e-6. $\lambda_\text{IFC}$ is set as 1 and $\lambda_{DPC}$ dynamically adapts to the $S_\text{IoU}$ scores. Additionally, the EMA rate of the teacher model is set to $0.95$. All experiments were conducted on a single NVIDIA GeForce 3090 GPU, implemented in PyTorch 2.5.1, to ensure a fair comparison.

\subsection{Quantitative Evaluations}
\subsubsection{Adaptation to Pancreas and Pancreatic Cancer Segmentation Tasks}
To assess in-distribution versus out-of-distribution (OOD) performance, we evaluate the Pancreas task from AMOS 2022~\cite{ji2022amos}, which was used to train MedSAM~\cite{ma2024segment} and SAM-Med2D~\cite{cheng2023sam}, and the Pancreatic cancer task from MSWAL~\cite{wu2025mswal}, which was published in 2025 and was not included in the training of these fine-tuned models.
As shown in Table~\ref{tab_sota_gray}, on the in-distribution Pancreas task the original SAM attains 63.79\% DSC. Fine-tuned models improve this (MedSAM: 67.71\%; SAM-Med2D: 79.36\%), confirming the benefit of supervised adaptation when training and testing are aligned. Among TTA methods, SAM-TTA achieves the best DSC at 74.41\%, surpassing all other TTA baselines (63.81\% to 70.64\%) and narrowing the gap to SAM-Med2D. SAM-TTA also obtains the lowest HD95 (14.26) compared with SAM (19.60) and MedSAM (15.36).
On the OOD Pancreatic cancer task, the fine-tuned models generalize poorly: MedSAM and SAM-Med2D drop to 69.50\% and 73.68\% DSC, both below the original SAM (81.57\%). In contrast, TTA methods consistently improve over SAM (for example, MT: 83.71\%, WESAM: 84.25\%), and our SAM-TTA attains the highest DSC of 84.68\%. SAM-TTA also achieves the best HD95 (4.92), outperforming SAM (6.12) and substantially reducing the large errors observed for MedSAM (11.34) and SAM-Med2D (16.28).
These results indicate that while fine-tuned models can excel on in-distribution data, it may degrade under distribution shift. By contrast, TTA, and particularly our SAM-TTA, delivers strong and stable gains on both overlap and boundary metrics without additional supervision.

\begin{table*}[htbp]
  \centering\caption{Ablation on brain tumor segmentation. Symbols follow Sec.~\ref{sec:method}. Here, $\diamond$ indicates a setting where, for each test image, the three channels are generated by per-image random cubic Bézier curves without any parameter learning.}

  \resizebox{1.0\linewidth}{!}{
    \begin{tabular}{cccccc|cccccccc}
    \toprule
    \multicolumn{6}{c|}{Our Designs in SAM-TTA}          & \multicolumn{2}{c}{BraTS-SSA T2W} & \multicolumn{2}{c}{BraTS-SSA T2F} & \multicolumn{2}{c}{BraTS-PED T2W} & \multicolumn{2}{c}{BraTS-PED T2F} \bigstrut\\
    \hline
    SAM   & SBCT & $\mathcal{L}_{\text{ICM}}$ &  $\mathcal{L}_{\text{DPC}}$ & Weight $\lambda_\text{DPC}$& $\mathcal{L}_\text{IFC}$ & DSC   & HD95  & DSC   & HD95  & DSC   & HD95  & DSC   & HD95 \bigstrut\\
    \hline
    \checkmark    &       &       &       &  &      & 78.77  & 25.06  & 79.77  & 23.49  & 83.62 & 15.71 & 85.67 & 13.38 \bigstrut[t]\\
    \midrule
    \checkmark    & \checkmark  & \checkmark      &       &  &   & 82.34 & 20.58 & 84.16 & 16.99 & 87.01 & 12.85 & 88.55 & 10.17 \\
    \checkmark    & \checkmark  & \checkmark      &\checkmark       &  &      &83.24     &20.94    &87.74  &13.07    & 89.06 &9.93   &90.59  &8.22  \\
    \checkmark   & \checkmark   & \checkmark      & \checkmark   & \checkmark   &       & 83.83  & 18.93  & 88.45  & 11.90  & 89.17  & 9.53  & 89.19  & 9.18  \\
    \checkmark   & \checkmark   & \checkmark   & \checkmark   & \checkmark  & \checkmark & 83.52  & 20.15  & 88.66  & 11.45  & 89.28  & 9.58  & 90.98  & 7.26  \\
    \midrule
     \checkmark  &    & \checkmark   & \checkmark   & \checkmark  & \checkmark & 80.13  & 26.71  & 81.84  & 23.67  & 88.39  & 11.29  & 89.84  & 9.10  \\
  \checkmark  &  $\diamond$  & \checkmark   & \checkmark   & \checkmark  & \checkmark & 80.03  & 26.93  & 83.72  & 19.93  & 88.70  & 10.61  & 90.49  & 7.97  \\
    \bottomrule
    \end{tabular}%
    }
  \label{tab_ablation}%
\end{table*}%

\subsubsection{Adaptation to Brain Tumor Segmentation}  
We further evaluate our model on the SSA and PED subsets of BraTS 2023. Although MedSAM~\cite{ma2024segment} and SAM-Med2D~\cite{cheng2023sam} are trained on the original BraTS dataset, these two subsets are excluded from their training. The task remains tumor segmentation, but the cohorts differ, which creates a principled test of generalization beyond the training distribution.
As shown in Table~\ref{tab_sota_gray}, the original SAM~\cite{kirillov2023segment} attains 78.77\% and 79.77\% DSC on BraTS-SSA (T2W and T2F), and 83.62\% and 85.67\% on BraTS-PED (T2W and T2F). SAM-Med2D performs worse than SAM on all four subsets (72.85\%, 79.32\%, 77.09\%, 84.87\%). MedSAM improves over SAM on each subset (82.70\%, 85.33\%, 87.38\%, 90.05\%). Our SAM-TTA achieves the highest DSC on all four subsets (83.52\%, 88.66\%, 89.28\%, 90.98\%).
Considering boundary accuracy, SAM-TTA also yields the lowest HD95 on all four subsets (20.15, 11.45, 9.58, 7.26), improving over MedSAM (19.50, 15.86, 11.70, 8.08) and the original SAM (25.06, 23.49, 15.71, 13.38). Averaged over the four brain tumor subsets, SAM-TTA reaches 88.11\% DSC and 12.11 HD95, compared with 86.36\% and 13.79 for MedSAM, and 81.96\% and 19.41 for SAM.
Results also show that entropy-minimization alone can be brittle under domain shift, whereas consistency-based TTA methods are more stable. For example, TENT~\cite{wang2021tent} exhibits large drops on BraTS-SSA (49.76\% DSC on T2W). Among all TTA baselines, our SAM-TTA is consistently the best in both overlap and boundary metrics on every subset.

\subsubsection{Adaptation to Polyp Segmentation Task}
Although our approach is primarily designed for grayscale medical images, we further assess its effectiveness on two RGB polyp benchmarks, CVC-ColonDB, which is used to train SAM-Med2D, and Kvasir-SEG, which is included in the training data of both fine-tuned references. As shown in Table~\ref{tab_sota_RGB}, the original SAM attains 87.58\% DSC and 8.20 HD95 on CVC-ColonDB, and 80.95\% DSC and 14.93 HD95 on Kvasir-SEG. Existing TTA baselines bring only modest gains on CVC-ColonDB, with DSC between 87.60\% and 87.97\%. Our SAM-TTA achieves 89.08\% DSC and 7.28 HD95, clearly improving over SAM and all TTA baselines.
On Kvasir-SEG, MT reaches 83.58\% DSC and 13.92 HD95, while SAM-TTA further improves to 85.55\% DSC and 13.15 HD95, outperforming all other TTA methods on both overlap and boundary metrics. Supervised fine-tuning references set a strong performance bar on these datasets: SAM-Med2D reaches 90.34\% DSC and 6.57 HD95 on CVC-ColonDB and 92.49\% DSC and 8.38 HD95 on Kvasir-SEG, and MedSAM attains 90.27\% DSC and 5.72 HD95 on CVC-ColonDB and 89.43\% DSC and 8.49 HD95 on Kvasir-SEG. Notably, without additional training or labels, SAM-TTA narrows much of the gap to these supervised references while preserving test-time adaptability.

\subsubsection{Qualitative Visualization}  
Fig.~\ref{fig_seg_sota} presents qualitative segmentation results across multiple datasets. 
The pretrained SAM often fails under distribution shifts, either producing fragmented masks (Pancreatic cancer), severe over-segmentation (BraTS-SSA), or missing fine structures (BraTS-PED). 
While existing TTA methods provide partial improvements, they are prone to unstable adaptation and residual artifacts, such as box-biased masks or excessive background leakage. 
MedSAM achieves better results for in-distribution data but degrade on unseen domains, as evidenced by BraTS subsets. 
In contrast, our method consistently delivers sharper boundaries and more complete structures across both grayscale and RGB modalities, producing results that closely match the ground truth and remain robust under diverse domain shifts.

\subsubsection{Ablation Analysis}
As shown in Table~\ref{tab_ablation}, adding SBCT{+}$\mathcal{L}_{\text{ICM}}$ to SAM yields the largest single-step gain across all four subsets: mean DSC rises from 81.96\% to 85.52\% (+3.56) while HD95 drops from 19.41 to 15.15 ($-4.26$). 
Per-subset trends are consistent (e.g., SSA~T2F: +4.39 DSC, $-6.50$ HD95; PED~T2F: +2.88 DSC, $-3.21$ HD95), indicating that input-level channel diversification via SBCT, together with IoU-guided certainty maximization, substantially contributes to better performance.
Enabling the mean-teacher and dual-scale consistency $L_{\text{DPC}}$ brings further improvements (85.52\%$\rightarrow$87.66\% DSC; 15.15$\rightarrow$13.04 HD95). 
Introducing IoU-weighted consistency ($w$) stabilizes adaptation and primarily benefits boundaries (HD95: 13.04$\rightarrow$12.39 on average), suggesting that the predicted IoU serves as a reliable confidence signal for test-time consistency.
Finally, adding $L_{\mathrm{IFC}}$ (IoU-tempered feature consistency) achieves the best overall results, reaching 88.11\% mean DSC and 12.11 mean HD95, with the largest boost on SSA~T2F (88.66\% DSC; 11.45 HD95).
To further isolate the effect of SBCT, we report two controls: (i) without SBCT but with MT+w+$\mathcal{L}_{\mathrm{DPC}}$+$\mathcal{L}_{\mathrm{IFC}}$ (mean DSC 85.05, HD95 17.69), and (ii) $\diamond$, where each test image’s three channels are generated by randomly sampled cubic Bézier curves with no parameter learning (mean DSC 85.74, HD95 16.36). Both trail our learnable SBCT configuration (mean DSC 88.11, HD95 12.11). This shows that simple channel diversification, whether absent or random, is insufficient, whereas adaptive and learnable intensity mappings are essential for performance gain.

\begin{figure*}
    \centering
    \includegraphics[width=0.98\linewidth]{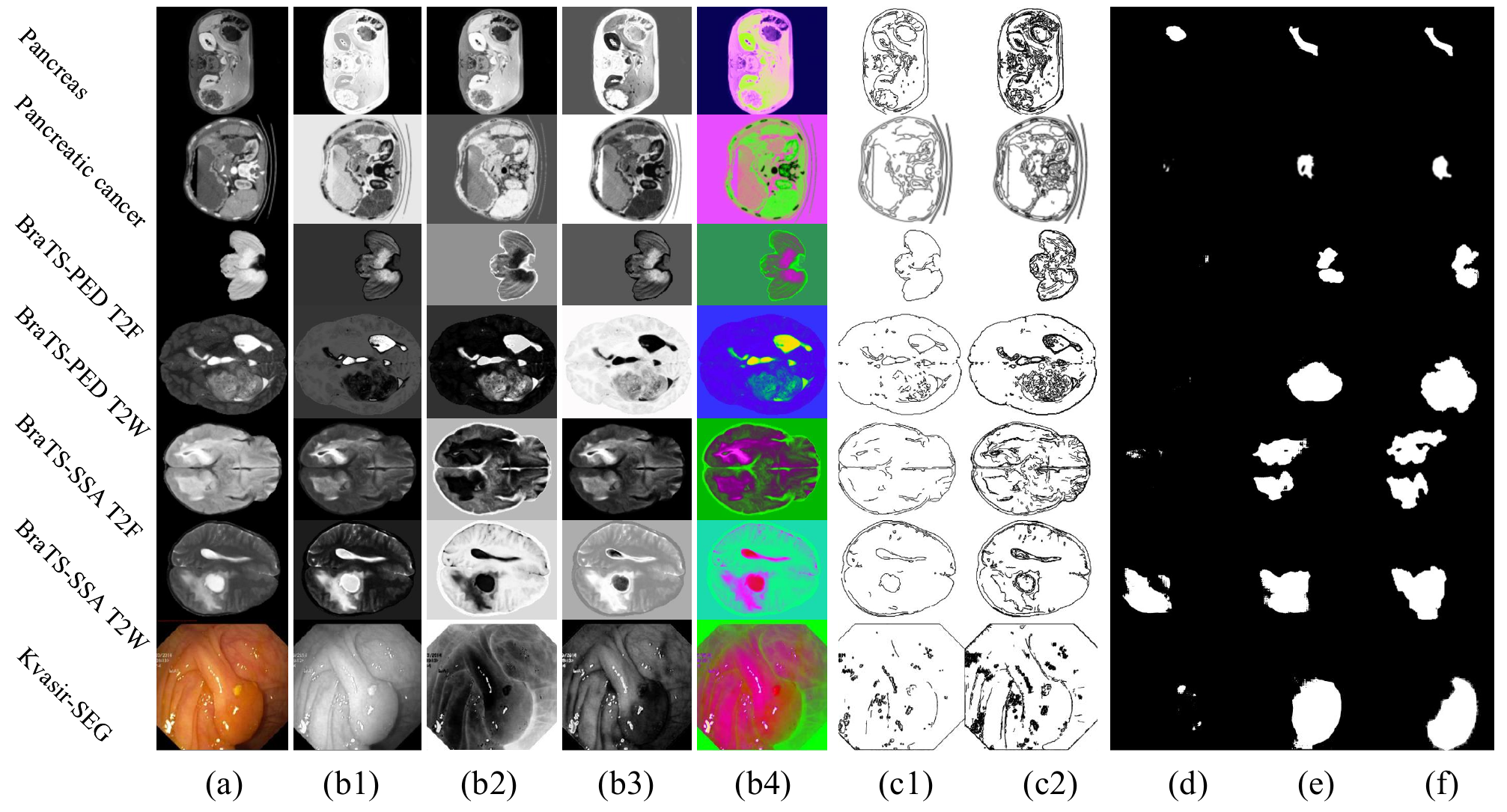}
\caption{Qualitative visualization of the proposed SBCT across seven datasets. (a) Original input slices. (b1–b3) Three SBCT-generated channels showing self-adaptive intensity remapping from a single-channel input. (b4) Pseudocolor composite of the three channels. (c1–c2) Canny edges extracted from (a) and (b4), respectively. (d) SAM predictions without adaptation. (e) Results after applying SBCT. (f) Ground truth annotations. SBCT enhances structural contrast and boundary definition, facilitating more accurate segmentation across both grayscale and color medical modalities.}
    \label{fig_SBCT_show}
    \vskip -0.4cm
\end{figure*}

\begin{figure}
    \centering
    \includegraphics[width=\linewidth]{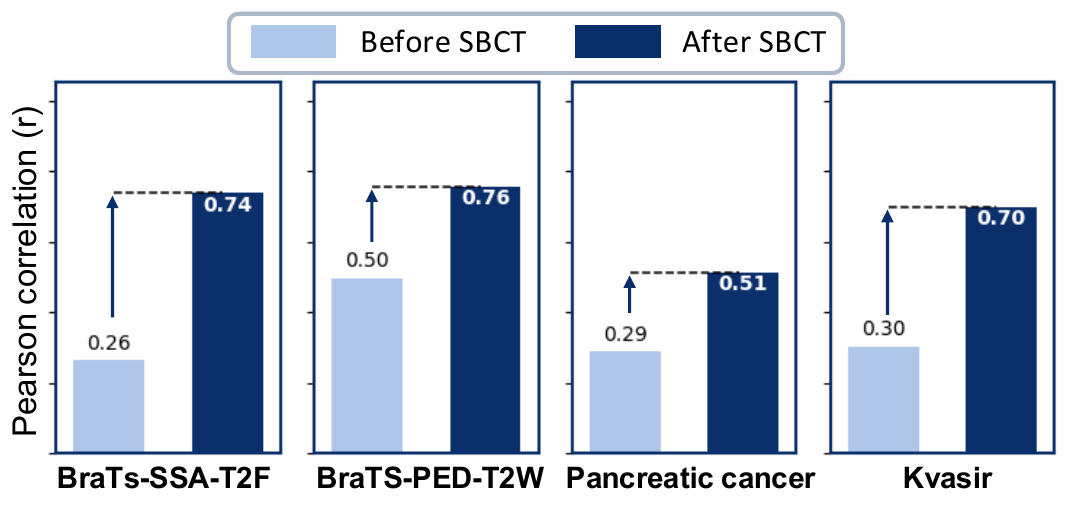}
    \caption{\textbf{Calibration of SAM's IoU predictor with and without SBCT.} 
We evaluate the Pearson correlation ($r$) between SAM's predicted IoU ($S_{\mathrm{IoU}}$) and the true IoU computed from the predicted high-resolution masks. 
Here SAM is kept \emph{frozen}, and only the SBCT parameters for generating the three-channel inputs are optimized.}
    \label{fig:sbct_calibration}
\end{figure}

\subsection{Analysis and Discussion}
\subsubsection{Calibration of SAM's IoU predictor with and without SBCT} 
As shown in Fig.~\ref{fig:sbct_calibration}, we further evaluate how SBCT affects the calibration of SAM’s intrinsic IoU predictor. In this experiment, SAM is kept completely \emph{frozen}, and only the SBCT parameters are optimized to generate the three-channel inputs. We measure the Pearson correlation ($r$) between SAM’s predicted IoU scores ($S_{\mathrm{IoU}}$) and the true IoU computed from the predicted high-resolution masks. SBCT consistently improves the correlation across all four datasets: BraTS-SSA-T2F ($0.26{\rightarrow}0.74$), BraTS-PED-T2W ($0.50{\rightarrow}0.76$), Pancreatic cancer ($0.29{\rightarrow}0.51$), and Kvasir ($0.30{\rightarrow}0.70$). These results indicate that SBCT yields better-calibrated uncertainty signals, which are crucial for reliable test-time adaptation.

\subsubsection{Boundary Enhancement through SBCT Inputs}
As illustrated in Fig.~\ref{fig_SBCT_show}, SBCT generates three complementary channels that amplify boundary-sensitive cues beyond simple grayscale replication. By adapting pixel intensities through Bézier curve remapping, SBCT enhances local contrast variations and preserves fine structural gradients. The resulting inputs reveal stronger and more continuous edge evidence (Fig.~\ref{fig_SBCT_show}c1–c2), especially around thin structures and low-contrast regions. When combined with our uncertainty-guided alignment, this leads to segmentation masks with sharper contours, reduced staircase artifacts, and substantially less background leakage (Fig.~\ref{fig_SBCT_show}e vs.\ d), yielding cleaner and more topologically consistent results across diverse datasets.

\subsubsection{Computational Efficiency}  
Our method outperforms MedSAM by requiring only a single optimization per image instead of time-consuming optimizations across numerous images. In addition, while WESAM and UPL-TTA require adaptation times of 0.58\,s per image and 0.60\,s per image, respectively, our method takes only 0.364\,s per image. Given that SAM’s direct inference requires 0.10\,s per image, our approach incurs an extra 0.26\,s per image, which is considered a small overhead for the substantial performance gains achieved.

\subsubsection{Limitation and Future Work}
Our study has three main limitations. First, we use oracle bounding-box prompts derived from ground truth, consistent with MedSAM, to ensure fair and reproducible comparisons and to avoid the instability of point prompts in low-contrast or ambiguous boundaries reported by prior work~\cite{ma2024segment}. However, such oracle boxes are rarely available in practice. Second, the backbone is 2D while many clinical scans are 3D, so volumetric context is not exploited and truly 3D structures may be suboptimally handled. Third, performance is bounded by SAM’s pretrained representations; in domains where the backbone struggles (for example, very low-contrast findings in non-contrast CT), test-time adaptation can offer only limited gains. Future work will explore extending our framework to 3D medical image segmentation and integrating additional self-supervised cues to improve adaptation.

\section{Conclusion}
We have presented SAM-TTA, a new test-time adaptation framework that enhances SAM for universal medical image segmentation by jointly addressing input- and semantic-level gaps. Specifically, the proposed SBCT converts single-channel scans into SAM-compatible three-channel inputs with few learnable parameters, effectively reducing input mismatches while preserving anatomical structures. In parallel, the designed IMA leverages SAM’s intrinsic IoU prediction to improve confidence, enforce dual-scale consistency, and align intermediate features between a LoRA-updated student and an EMA teacher, while keeping the mask decoder frozen to retain IoU calibration. Extensive experiments on eight datasets spanning MRI, CT, and endoscopy demonstrate that SAM-TTA consistently outperforms state-of-the-art TTA methods and, in several out-of-distribution settings, even matches or surpasses fully fine-tuned baselines such as MedSAM and SAM-Med2D, without retraining or labeled data.

\bibliographystyle{IEEEtran}
\bibliography{myref}
\end{document}